\documentclass[letterpaper]{article} 
\usepackage{aaai24}  
\usepackage{times}  
\usepackage{helvet}  
\usepackage{courier}  
\usepackage[hyphens]{url}  
\usepackage{graphicx} 
\usepackage{natbib}  
\usepackage{caption} 
\usepackage{algorithm}
\usepackage{algorithmic}

\usepackage{newfloat}
\usepackage{listings}
\DeclareCaptionStyle{ruled}{labelfont=normalfont,labelsep=colon,strut=off} 
\floatstyle{ruled}
\newfloat{listing}{tb}{lst}{}
\floatname{listing}{Listing}
\usepackage{booktabs}       
\usepackage{amsfonts}       
\usepackage{nicefrac}       
\usepackage{microtype}      
\usepackage{xcolor}         
\usepackage{subcaption}  
\usepackage{amsmath}
\usepackage{amssymb}
\usepackage{mathtools}
\usepackage{amsthm}
\usepackage[textsize=tiny]{todonotes} 
\usepackage[capitalize,noabbrev]{cleveref}

\DeclareMathOperator*{\argmax}{arg\,max}
\usepackage{enumitem}
\usepackage{dsfont}
\usepackage{makecell}

\title{Consistency-Guided Temperature Scaling Using Style and Content Information \\ for Out-of-Domain Calibration}
\author{
    Wonjeong Choi\textsuperscript{\rm 1},
    Jungwuk Park\textsuperscript{\rm 1},
    Dong-Jun Han\textsuperscript{\rm 2},
    Younghyun Park\textsuperscript{\rm 1},
    Jaekyun Moon\textsuperscript{\rm 1}
}
\affiliations{
    \textsuperscript{\rm 1}Korea Advanced Institute of Science and Technology (KAIST) \\
    \textsuperscript{\rm 2}Purdue University \\
  
    \{dnjswjd5457, savertm, dnffkf369\}@kaist.ac.kr, han762@purdue.edu, jmoon@kaist.edu
    
}

\begin{document}

\maketitle

\begin{abstract}
	Research interests in the robustness of deep neural networks  against domain shifts have been rapidly increasing in recent years. Most existing works, however, focus on improving the accuracy of the model, not the calibration performance which is another important requirement for trustworthy AI systems. Temperature scaling (TS), an accuracy-preserving post-hoc calibration method, has been proven to be effective in in-domain settings, but not in out-of-domain (OOD) due to the difficulty in obtaining a validation set for the unseen domain beforehand. In this paper, we propose consistency-guided temperature scaling (CTS), a new temperature scaling strategy that can significantly enhance the OOD calibration performance by providing mutual supervision among data samples in the source domains. Motivated by our observation that over-confidence stemming from inconsistent sample predictions  is the main obstacle to OOD calibration, we propose to guide the scaling process by taking consistencies into account in terms of two different aspects - style and content - which are the key components that can well-represent data samples in multi-domain settings. Experimental results demonstrate that our proposed strategy outperforms existing works, achieving superior OOD calibration performance on various datasets. This can be accomplished by employing only the source domains without compromising accuracy, making our scheme directly applicable to various trustworthy AI systems. 
\end{abstract}

\section{Introduction}
Despite the huge success of deep neural networks (DNNs) in various fields  such as computer vision \cite{krizhevsky2017imagenet} and natural language processing \cite{mikolov2013efficient}, it is still difficult to actively utilize  DNNs in safety-critical or high-risk applications including medical engineering and defect detection.  One of the main reasons is that there is a huge risk when the prediction of the model becomes incorrect. In such applications, it is important for the neural network  to   reliably indicate whether its prediction is likely to be  correct  or not.  In other words, the model should have a reliable confidence about its prediction. Given the model that has a reliable   confidence,  we can use the prediction of the model (e.g., for disease diagnosis) if the prediction is likely to be correct  (i.e.,  high confidence),  but    may rely more on other factors (e.g., decision of the doctor) if  the prediction is likely to be incorrect  (i.e., low confidence).  Therefore, promoting the quality of the prediction confidence of DNNs is a key mission for trustworthy AI \cite{tomani2021towards, guo2017calibration}.  However,  the problem of modern DNNs is that they often produce over-confident predictions, which means that the confidence of predictions is not reliable.

In this context, model calibration has been     developed  as an important research direction      to improve the  reliability of   confidence of DNNs.  In general, we say that the model is well-calibrated if the  confidence  approximates the probability of being correct well. The authors of \cite{guo2017calibration} have shown that modern neural networks have poor calibration performance due to over-confident predictions, and suggested a simple yet effective   method termed temperature scaling (TS).  TS  is a post-hoc calibration method that calibrates the confidence of the trained model to make the confidence similar to a true probability using a validation set, without affecting the performance of the trained model.  Various other methods such as train-time regularization \cite{thulasidasan2019mixup, krishnan2020improving, mukhoti2020calibrating, hebbalaguppe2022stitch} and probabilistic approaches \cite{gal2016dropout, lakshminarayanan2017simple} have been also proposed in the literature improve the model's calibration performance.

However, most prior works   focus   on the  in-domain  setting, under the assumption that the domain distributions of train data and test data are the same. This assumption often does not hold in practice having domain shifts between training and testing: in out-of-domain (OOD) settings, the neural network needs to make predictions for samples from the unseen domain (target domain) that has not been observed in the training set (source domains).   
Existing calibration strategies face great challenges in the OOD settings due to the difficulty of knowing the unseen domain beforehand. 
Although several works \cite{tomani2021post, gong2021confidence, yu2022robust} have recently suggested some TS-based calibration methods under domain shift, their calibration performance is still limited when the disparity between the source and target domains becomes severe, as we will see   in Section  \ref{experiments}.


\paragraph{Main contributions.} 
To handle the fundamental challenges of existing calibration methods in OOD settings, we propose consistency-guided temperature scaling (CTS), a new TS-based post-hoc strategy that can achieve superior OOD calibration performance. Motivated by the intuition that keeping the consistency of sample predictions (regardless of domain shifts) can enhance the reliability of the model prediction in the unseen domain, our CTS trains a scaling temperature tailored to OOD scenarios, which enables the calibrated model to produce domain-invariant predictions. Our core idea is to take consistencies into account during temperature scaling, in terms of two different aspects: style and content, which are the key components representing data samples in multi-domain settings. To optimize a temperature, we compose a new TS objective function that includes two auxiliary losses for improving 1) inter-domain style consistency and 2) intra-class content consistency, thereby considering both aspects to achieve a better OOD calibration.

To gain insights into our idea and to support our claim, we start by analyzing the correlation between the two consistencies and OOD calibration, and show that   the model's inconsistent predictions (on  style and content variations) cause over-confident predictions on the target domain, leading to poor OOD calibration. Our CTS is designed to tackle this issue in OOD settings by  optimizing  the temperature to make   predictions that are invariant to style and content shifts in the target domain.  Surprisingly, our approach can significantly improve the OOD calibration performance (while preserving the accuracy) by strategically optimizing the consistency-guided temperature using style and content information only from the source domains. 
This is a key advantage  of CTS  as the temperature can be successfully optimized by utilizing the useful attributes in the source domains, without requiring any  target domain information.


Experimental results on various multi-domain datasets   show that   CTS can achieve remarkable OOD calibration performance  compared to  existing TS-based methods.  Notably, the performance gain of CTS becomes larger when the domain disparity is large, supporting the effectiveness of our method in practical OOD settings.


\section{Related Works} \label{rw}
\paragraph{Calibration methods.} 
A large body of literature has been devoted to improving the calibration performance of deep neural networks in in-domain settings.
One of research directions is a train-time calibration \cite{thulasidasan2019mixup, krishnan2020improving, mukhoti2020calibrating, hebbalaguppe2022stitch, liu2022devil}, which prevents over-confident predictions by regularizing the model during training to enhance calibration performance. Another line of works is a post-hoc method \cite{guo2017calibration, vovk2005algorithmic, zadrozny2002transforming, zadrozny2001obtaining}, which adjusts the confidence of the model output after training. Especially, temperature scaling (TS) \cite{guo2017calibration} has been widely used since it can be easily combined with any trained model without compromising the original accuracy. TS prevents over-confident or under-confident predictions for test data by scaling the output with a single parameter (temperature) optimized on the validation set. However, although TS has been proven to be effective in in-domain settings, achieving a good calibration on OOD samples remains a great challenge due to the difficulty in obtaining a validation set for the unseen domain beforehand.

\paragraph{Calibration for out-of-domain (OOD) scenarios.} 
Only a few works have proposed TS-based methods for OOD calibration. 
\cite{tomani2021post} generates an augmented validation set which can simulate domain shift by injecting perturbations into validation samples before the post-hoc calibration. However, this scheme highly depends on the perturbation degree of the noise. 
The authors of \cite{gong2021confidence} utilize multiple domains to reduce the distributional gap between the target and calibration domains for improved calibration transfer. 
Specifically, after clustering the calibration domains, the model is calibrated with the temperature of a specific group, which is most similar to the unseen domain encountered during the OOD inference.
Similarly, \cite{yu2022robust} leverages multiple domains (with domain labels) to train a linear regression model which predicts sample-wise temperature for the target domain. 
However, the performances of these works  \cite{gong2021confidence, yu2022robust} are still limited especially when the domain discrepancy between the calibration domains and the target domain is large. 


Compared to prior works, we take advantage of style and content information in the source domains and effectively optimize a temperature to enable the calibrated model to make domain-invariant predictions regardless of domain shifts, achieving superior OOD calibration performance even with a large domain gap between training and inference. 

\paragraph{Domain generalization (DG).}
The goal of DG \cite{gulrajani2020search, zhou2021domain} is to improve the generalization ability of DNNs, that is, models trained from the source domain work well even in the unseen target domain. 
The domain-alignment-based DG \cite{muandet2013domain, motiian2017unified, li2018domain} aims to learn features invariant to the domain shifts in the feature space through a domain-invariant learning. 
The augmentation-based method \cite{zhou2021mixstyle, li2022uncertainty, shankar2018generalizing} prevents the model from over-fitting to the source domains by using augmented data which simulate domain shift during training. Although existing DG methods have recently achieved high accuracy in unseen domains, one major bottleneck hindering their practical usability is a lack of calibration capability \cite{gong2021confidence}. In this paper, we present an effective calibration method that  facilitates trustworthy AI systems in multi-domain settings by dealing with the calibration aspect for OOD scenarios.

\begin{figure*}[t]
	\centering
	\begin{subfigure}[b]{0.245\textwidth}
		\centering	\includegraphics[width=\textwidth]{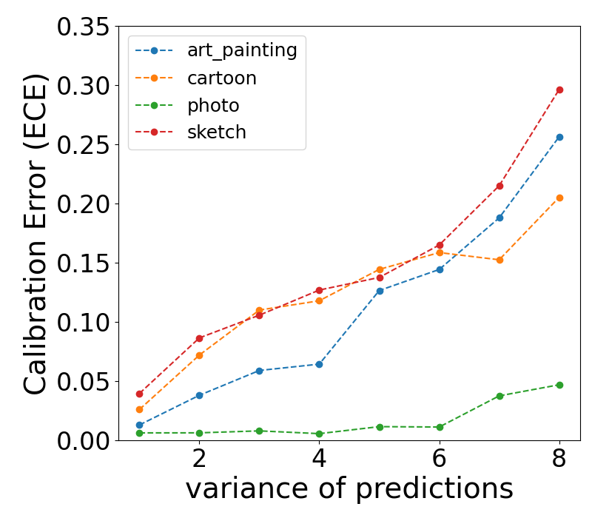}
		\caption{After style shifts}
		\label{fig:scons_vs_ece}    
	\end{subfigure}
	\hspace{-2mm}
	\begin{subfigure}[b]{0.245\textwidth}
		\centering
		\includegraphics[width=\textwidth]{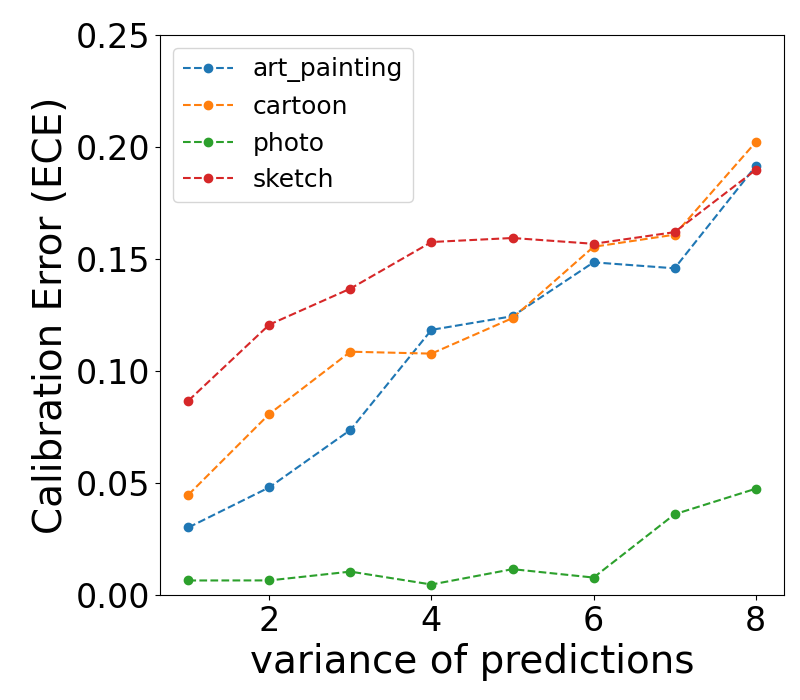}
		\caption{After content shifts}
		\label{fig:ccons_vs_ece}    
	\end{subfigure}
	\hspace{+2mm}
	\begin{subfigure}[b]{0.235\textwidth}
		\centering
		\includegraphics[width=\textwidth]{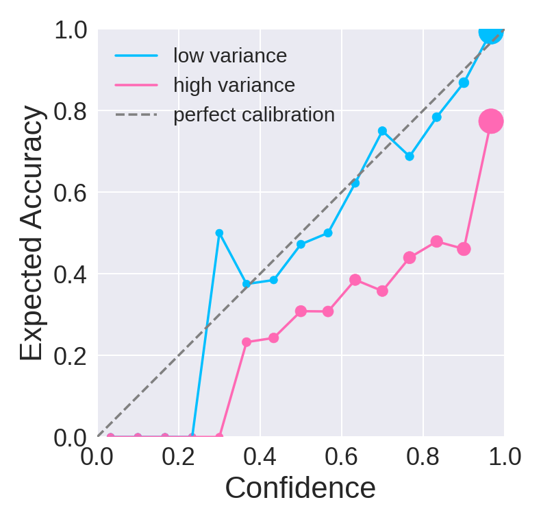}
		\caption{Style perspective}
		\label{fig:style_relidiag}    
	\end{subfigure}
	\hspace{-3mm}
	\begin{subfigure}[b]{0.235\textwidth}
		\centering
		\includegraphics[width=\textwidth]{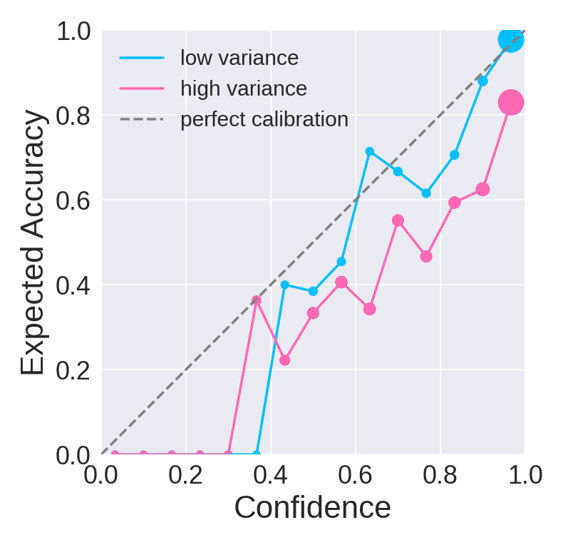}
		\caption{Content perspective}
		\label{fig:content_relidiag}    
	\end{subfigure}
	\caption{(a-b) Correlations between variance   of predictions and OOD calibration performance for test samples from different target domains. Samples with high variance  are more likely to show poor OOD calibration performance in both cases of style and content shifts on PACS dataset. (c-d) Reliability diagrams for comparing calibration tendency depending on the variance of predictions under style and content variations. It can be confirmed that the poor calibration performance of high variance samples (pink line) arises from the over-confident predictions. We note that the size of points indicates the relative counts of samples in each corresponding confidence interval.}
	\label{fig:correlation} 
\end{figure*}

\section{Background}

\subsection{Temperature Scaling (TS)} 
As an accuracy-preserving approach that calibrates a pre-trained model in a post-hoc manner, TS \cite{guo2017calibration} is widely applied in real applications due to its simplicity and effectiveness. TS aims to prevent the model from making over- or under-confident predictions by adjusting a single temperature $T$, without updating the parameters of the pre-trained model.
The temperature is optimized by utilizing a validation dataset to minimize the negative log-likelihood (NLL) loss $\mathit{L}_{NLL}$ in the Eq. (\ref{eq1}), using temperature scaled confidence score $\mathbb{P}(y=y_i|f_i, T) = \frac{\exp({f_{i,y_i}/T})}{\sum_{k=1}^K \exp({f_{i,k}/T})}$ for ground truth $y_i$, where $f_{i}$ is a logit vector of the sample $x_i$ and $f_{i,k}$ is $k$-th element of $f_{i}$ for class $k$. The optimized temperature $T^*$ can be defined as


\begin{equation}
	\begin{aligned} \label{eq1}
		T^* &= \operatorname*{arg\,min}_T 
		\left [
		\frac{1}{N_{val}}
		\sum_{i=1}^{N_{val}} 
		\mathit{L}_{NLL}
		\right ] \\
		&= \operatorname*{arg\,min}_T \left [
		\frac{1}{N_{val}}
		\sum_{i=1}^{N_{val}} 
		-y_i\log (\mathbb{P}(y=y_i|f_i, T))
		\right ],
	\end{aligned}
\end{equation} 
where $N_{val}$ is the number of samples in the validation set. Conventional TS works well in in-domain scenarios where the data distributions between the validation set (for calibration) and test set are identical. 
However, in OOD scenarios, the model should be guided to make calibrated predictions on target domains unseen during training or calibration, which is a great challenge in the presence of  domain shift \cite{tomani2021post, gong2021confidence, yu2022robust}.

\subsection{Style Shifting} \label{style_and_content_shifting}
It is well known that feature statistics at early CNN layers can represent domain/style information of each data sample \cite{huang2017arbitrary}. Specifically, given a sample $x_i$, the intermediate CNN feature $z_i \in \mathbb{R}^{C\times H\times W}$ of $x_i$  at a specific layer can be interpreted as
\begin{equation}\label{feature_eq}
	\begin{aligned}
		z_i = \underbrace{\sigma(z_i)}_{\text{style} } \cdot \underbrace{\frac{z_i-\mu(z_i)}{\sigma(z_i)}}_{\text{content}} + \underbrace{\mu(z_i)}_{\text{style}},
	\end{aligned}
\end{equation} where $\mu(z_i) = \frac{1}{HW}\sum_{h=1}^H\sum_{w=1}^W z_{i,(c,h,w)}$ and $\sigma^2(z_i) = \frac{1}{HW}\sum_{h=1}^H\sum_{w=1}^W(z_{i,(c,h,w)} - \mu(z_i))^2$. Here, $C$, $H$, and $W$ denote the channel, height, and width of the feature maps, respectively. The feature statistics $\mu(z_i), \sigma(z_i)$ (also called style statistics) contain characteristics of each domain (i.e., style information) while content information is preserved in the style-normalized feature in Eq. \ref{feature_eq}. For notational simplicity, we define style $s_i$ and content $c_i$ of sample $x_i$ as $s_i:=(\mu(z_i), \sigma(z_i))$ and $c_i := \frac{z_i-\mu(z_i)}{\sigma(z_i)}$. Based on this observation, adaptive instance normalization (AdaIN) \cite{huang2017arbitrary} was  introduced to transfer the style of $x_i$ to a different style of another sample $x_j$ by replacing the feature statistics from $s_i$ to $s_j$.


\section{Consistency-Guided Temperature Scaling}

\subsection{Problem Setup}

We consider a multi-class classification task with $K$ classes. Let $x_i$, $y_i$ denote the $i$-th image sample and the corresponding label drawn from a joint data distribution $P(x, y)$. For the multi-domain scenario, we let $P^{\mathcal{S}_j}$ denote the distribution of  the $j$-th source domain and we have a total of $J$ domains ($P^{\mathcal{S}_1}, ... , P^{\mathcal{S}_J}$). Data from the source domains are separated into (i) \textit{training set} that is used for training the base model and (ii) \textit{validation set} utilized for post-hoc calibration. Given the model pre-trained on the training set, our goal is to learn a temperature $T$ that can well calibrate the model on an arbitrary target domain  (with distribution $P^\mathcal{T}$) using the validation set, without changing the parameters of the pre-trained model (i.e., in a post-hoc manner). Note that if the model is perfectly calibrated in OOD scenarios, the confidence score should perfectly reflect the accuracy regardless of domains  so that $\mathbb{P}(\hat{y}=y|\hat{p}_x)=\hat{p}_x$ holds for each sample $(x, y)$ drawn from any $P^\mathcal{T}$, where $\hat{y}$ and $\hat{p}_x$ are the predicted class and the confidence for sample $x$, respectively.

\begin{figure*}[t]
	\centering
	\includegraphics[width=0.8\textwidth]{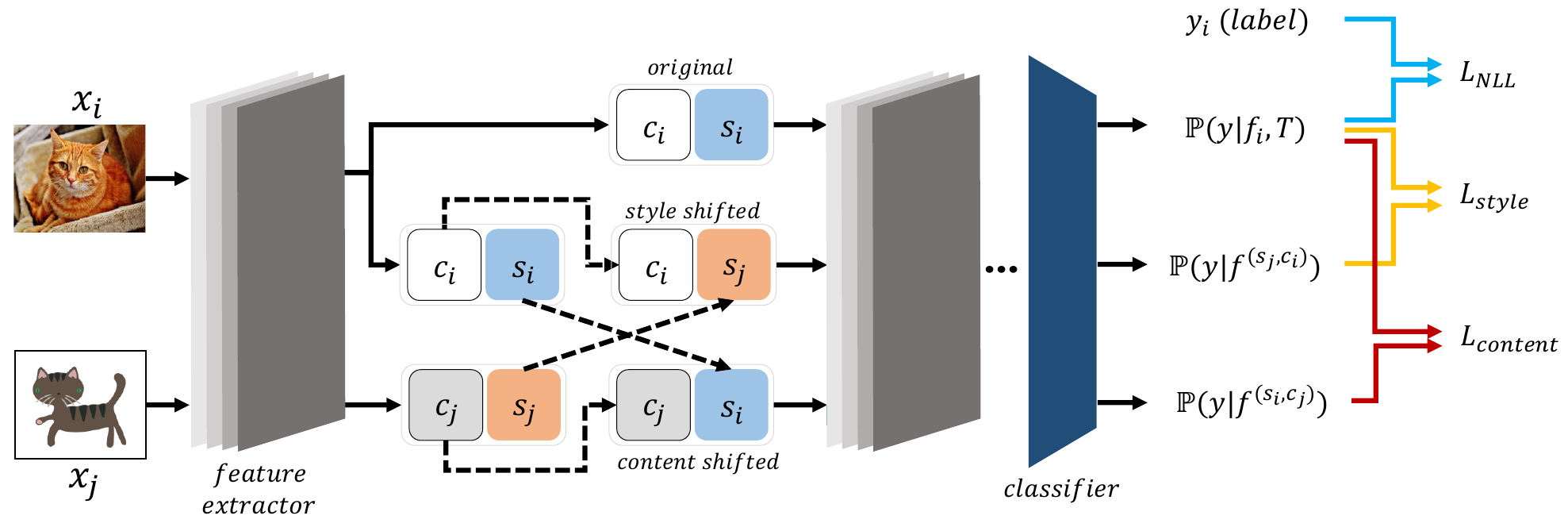} 
	\caption{Overview of our consistency-guided temperature scaling (CTS). Samples from the same class on the validation set are fed into the model in a pair-wise manner, and three different intermediate features (original, style shifted, content shifted) are generated. Then, style/content shifted logits {\small $\mathbb{P}(y|f^{(s_j, c_i)})$/$\mathbb{P}(y|f^{(s_i, c_j)})$} are created, and TS  is performed with consistency losses   described in Section \ref{overallloss}.}
	\label{fig:overview}
\end{figure*}

\subsection{Key Insights: Correlations between Consistency and OOD Calibration} \label{insights}
Our intuition is that keeping the consistency of predictions under  style and content perturbations can improve the reliability of the model prediction on the unseen domain, resulting in good OOD calibration.  
To gain insights into our idea based on this intuition, we provide an analysis of the correlation between OOD calibration and style/content consistencies using PACS dataset with 4 domains (Photo, Art, Cartoon, Sketch). 

\paragraph{Setup.}To simulate the OOD scenarios, the base model is trained on three source domains and evaluated on the remaining one domain (target domain). Regarding the style consistency, given the base model pre-trained on the training set, we first obtain 4 different predictions for each test sample (in the target domain) by changing its style  to 4 different styles (3 randomly chosen styles from each domain in the validation set, and 1 for the original style) while maintaining the content of each sample via Eq. \ref{feature_eq}. For the content consistency, we generate 5 different predictions by injecting different levels of zero-mean Gaussian noises with different variance [0, 0.1, 0.2, 0.3, 0.4] into the content $c_i$ in Eq. \ref{feature_eq} of each test sample. After obtaining  multiple predictions in each case, an indicator for each sample's  style/content consistency is measured as follows: the variance of each element in the  logit  vector is first computed with  different predictions, and then we average these results to obtain a scalar value. 
Finally, we sort test data according to their measured style/content consistency (i.e., variance of predictions), and measure the average expected calibration error (ECE) among samples with the same consistency value. 

\paragraph{Key observations.} Based on these measurements, Figs. \ref{fig:scons_vs_ece}, \ref{fig:ccons_vs_ece} show the trend of ECE according to style/content consistency. It can be confirmed that the variance of predictions under style/content shifts ($x$-axis) is highly related to the calibration error ($y$-axis), which means that the higher the variance (i.e., the lower the consistency) under the style/content shifts, the higher the calibration error.

\paragraph{Causality analysis.}
To explore the cause-effect relation between calibration and the variance of predictions, we plot reliability diagrams \cite{niculescu2005predicting} of the top 5\% samples and the bottom 5\% samples in terms of variances of predictions after applying style/content shifts.
Note that the perfect calibration is the identity function ($y=x$) while lines below the diagonal indicate that the model tends to output over-confident predictions, and vice versa. In Figs. \ref{fig:style_relidiag}, \ref{fig:content_relidiag}, it can be observed that inconsistent predictions with high variance (pink) tend to be over-confident compared to the consistent predictions (blue) after both shifts. 
These results indicate that samples that are sensitive to style and content variations cause  unstable  model predictions, and lead to over-confidence  as shown  in Figs. \ref{fig:style_relidiag}, \ref{fig:content_relidiag}. 
Inspired by these empirical observations, we propose CTS to improve OOD calibration on arbitrary target domains by optimizing the temperature such that the predictions are more consistent across variations in style/content.

\subsection{Proposed CTS for OOD Calibration}
In this section, we introduce our method called consistency-guided temperature scaling (CTS). CTS strategically optimizes the temperature such that predictions are more consistent across the source domains. The core idea of CTS is to incorporate consistencies in two different aspects, style and content, which are independent components to represent data samples in multi-domain settings. We design new auxiliary cost functions for temperature optimization with consideration for style- and content-consistency to achieve better OOD calibration performance. We illustrate our CTS in Fig. \ref{fig:overview} and explain the details of components in the following sections.

\subsubsection{Inter-domain style consistency loss.}
The observations in Figs. \ref{fig:scons_vs_ece} and \ref{fig:style_relidiag} imply that consistency of predictions under style shift is closely related to OOD calibration. In particular, over-confident predictions tend to deteriorate calibration performance. Therefore, we aim to improve OOD calibration by rescaling predictions of the pre-trained model so that it can be consistent regardless of inter-domain style changes.  Contrary to previous works \cite{motiian2017unified, li2018domain, kang2019contrastive} that focused on  updating model parameters to learn style-invariant features during training, CTS optimizes a style-invariant temperature to recalibrate pre-trained models. In this context, using a validation set from source domains, we propose the style consistency loss $\mathit{L}_{style}$ as:

\begin{equation}
\begin{aligned} \label{styleloss}
\mathit{L}_{style} &= D_{KL}[\mathbb{P}(y\vert f_i, T) \parallel \mathbb{P}(y\vert \overbrace{%
	\vphantom{ 1.0 }f^{(s_j, c_i)}}^{\substack{\text{style-shifted}}})] \\
&= \sum_{k=1}^{K}\mathbb{P}(y=k|f_i, T) \log{\frac{\mathbb{P}(y=k|f_i, T)}{\mathbb{P}(y=k|f^{(s_j, c_i)})}},
\end{aligned}
\end{equation} 
where $D_{KL} [\cdot\parallel\cdot]$ is KL-divergence and $f^{(s_j, c_i)}$ is a style-shifted logit with style $s_i$ of sample $x_i$ substituted by $s_j$ of sample $x_j$.
By strategically utilizing style information of validation samples from source domains, CTS can simulate a variety of style variations while optimizing the domain-invariant temperature to make consistent predictions. Concretely, as illustrated in Fig. \ref{fig:overview}, for the sample pair ($x_i, x_j$), the corresponding intermediate features ($z_i$, $z_j$) are separated into ($s_i$, $c_i$) and ($s_j$, $c_j$) through the operation in Eq. \ref{feature_eq}. Then, the style $s_i$ of sample $x_i$ is replaced with $s_j$ of sample $x_j$ to create a style-shifted feature, and it goes through the remaining layers to generate a style-shifted logit $f^{(s_j, c_i)}$ along with the original logit $f_i$. Lastly, the distance between temperature-scaled output $\mathbb{P}(y|f_i, T)$ and style-shifted output $\mathbb{P}(y| f^{(s_j, c_i)})$ is minimized such that the model is calibrated to generate style-invariant predictions. We note $\mathit{L}_{style}$ focuses on inter-domain style changes by minimizing the divergence between model outputs from features that differ only in style aspect.

\subsubsection{Intra-class content consistency loss.}
Another important component that represents data samples in multi-domain settings is content: complex semantic features used to distinguish classes of samples. We note that even samples belonging to the same class can have different contents depending on some properties of the object (e.g., location, posture, shape, etc). As observed in Figs. \ref{fig:ccons_vs_ece} and \ref{fig:content_relidiag}, content consistency is proportionally correlated with OOD calibration performance, similar to the trend observed with style. Also, models tend to make over-confident predictions on target samples that are relatively sensitive to the variations in content (as shown in Fig. \ref{fig:content_relidiag}).  Therefore, it is important to penalize such inconsistent predictions across intra-class content variation. To this end, CTS incorporates intra-class content consistency loss $\mathit{L}_{content}$ into the temperature optimization process as: 

\begin{equation}
\begin{aligned} \label{contentloss}
\mathit{L}_{content}
&=D_{KL}[\mathbb{P}(y|f_i, T) \parallel \mathbb{P}(y|\overbrace{%
	\vphantom{ 1.0 }f^{(s_i, c_j)}}^{\substack{\text{content shifted}}})] \\
&= \sum_{k=1}^{K}\mathbb{P}(y=k|f_i, T) \log{\frac{\mathbb{P}(y=k|f_i,T)}{\mathbb{P}(y=k|f^{(s_i, c_j)})}},
\end{aligned}
\end{equation} 
where $f^{(s_i, c_j)}$ is a content-shifted logit with content $c_i$ of sample $x_i$ substituted by $c_j$ of sample $x_j$. Note that $x_j$ belongs to the same category as $x_i$. As shown in Fig. \ref{fig:overview}, our CTS minimizes the divergence (as in Eq. \ref{contentloss}) between the two predictions obtained from the original logit $f_i$ and the content-shifted logit $f^{(s_i, c_j)}$ while keeping style $s_i$ the same. After all, $\mathit{L}_{content}$ guides the model to make consistent predictions under the variation of contents.

\subsubsection{Overall loss of our CTS.} \label{overallloss}
CTS operates in a pair-wise manner; two validation samples in a batch from source domains provide mutual supervision to each other, to find a scaling temperature that improves consistency in terms of style and content. 
Without any  target domain information, CTS effectively takes advantage of the  style and content features in the source domains for temperature scaling. Consequently, CTS obtains consistency-guided temperature $T^*_{CTS}$ from a validation set as:
\begin{equation}
\begin{aligned} \label{totalloss}
&L_{total} = \mathit{L}_{NLL} + \lambda_1 \cdot \mathit{L}_{style} + \lambda_2 \cdot \mathit{L}_{content}, \\
&T^*_{CTS} = \operatorname*{arg\,min}_T 
\left [
\frac{1}{N_{val}}
\sum_{i=1}^{N_{val}} 
L_{total}
\right ]
\end{aligned}
\end{equation}
Note that $L_{total}$ combines NLL loss in Eq. \ref{eq1} with style and content consistency losses in Eq. \ref{styleloss} and Eq. \ref{contentloss}. Here, $\lambda_1$ and $\lambda_2$ are coefficients for each loss, which are used to adjust the balance of two losses. 
These coefficients can be  determined depending on each dataset's distributional characteristics by using its validation set,  where more details are provided in supplementary material \footnote{Our supplementary material can be found in the github page at \\ https://github.com/wonjeongchoi/CTS.git}. 


\section{Experiments} \label{experiments}

\begin{table*}[!t]
\centering
\begin{subtable}[!t]{1\linewidth}
\small
\centering
\begin{tabular}{c||cccc|c||cccc|c}
\toprule  
& \multicolumn{5}{c}{\textbf{PACS}} & \multicolumn{5}{c}{\textbf{VLCS}} \\
\cmidrule{2-11}
Methods   & Art & Cartoon & Photo & Sketch & Avg. & Caltech & LabelMe & Pascal & Sun & Avg.  \\
\midrule
Vanilla        			&11.07&12.32& 1.30&14.62& 9.83 $\pm$ 0.45& 2.31&32.01&12.68&20.71&16.93 $\pm$ 0.38\\	
TS with val             & 9.70&11.30& 1.01&14.21& 9.06 $\pm$ 0.47& 5.38&29.03& 7.01&14.57&14.00 $\pm$ 0.39\\
PerturbTS      			& 8.24& 6.33&15.70& 6.49& 9.19 $\pm$ 1.19&18.30&23.84& 9.04& 9.32&15.13 $\pm$ 0.91\\
MDTS		            &10.33&10.98& 1.49&15.18& 9.50 $\pm$ 0.88& 2.78&28.24& 7.37&13.94&13.08 $\pm$ 0.38\\
CCDG-NN				    & 9.92&10.82& 1.30&13.16& 8.80 $\pm$ 0.44& 5.29&28.44& 7.24&13.93&13.73 $\pm$ 0.43\\
CCDG-Reg		        & 9.14&11.18& 1.45&13.46& 8.81 $\pm$ 0.61& 4.97&29.65& 6.91&14.03&13.89 $\pm$ 0.51\\
CCDG-Ens		        & 9.89&11.33& 1.15&13.86& 9.06 $\pm$ 0.41& 4.69&29.21& 7.16&14.38&13.86 $\pm$ 0.43\\
\textbf{CTS (ours)}     & 3.20& 3.05& 8.59&	2.38& \textbf{4.31 $\pm$ 0.93}&11.30&24.09& 6.01&10.18& \textbf{12.90 $\pm$ 0.81}\\
\cmidrule{1-11}
TS with test      & 3.82& 5.34& 1.15& 4.07& 3.60 $\pm$ 0.24& 2.15&15.10& 6.96&10.53& 8.69 $\pm$0.19\\
\bottomrule
\end{tabular}
\end{subtable}
\hfill
\newline
\begin{subtable}[!t]{1\linewidth}
\small	
\centering
\begin{tabular}{c||cccc|c||cccc|c}
\toprule  
& \multicolumn{5}{c}{\textbf{Office-Home}} & \multicolumn{5}{c}{\textbf{Digits-DG}} \\
\cmidrule{2-11}
Methods   & Art & Clipart & Product & Real & Avg. & MNIST & MNIST-M & SVHN & SYN & Avg.  \\
\midrule
Vanilla        			&12.22&17.14& 6.27& 6.25&10.47 $\pm$ 0.21& 1.65&23.64&18.10& 6.79&12.55 $\pm$ 0.17\\	
TS with val             &11.41&16.47& 5.54& 5.75& 9.79 $\pm$ 0.19& 1.13&21.21&16.60& 5.38&11.08 $\pm$ 0.15\\
PerturbTS		        & 7.52&10.66& 2.96& 3.02& 6.04 $\pm$ 0.62& 7.88& 4.98& 4.68&11.73& 7.32 $\pm$ 0.93\\
MDTS		            &10.95&17.81& 6.46& 5.99&10.30 $\pm$ 0.42& 1.24&20.70&16.59& 4.85&10.85 $\pm$ 0.19\\
CCDG-NN			        &11.51&16.46& 5.47& 5.79& 9.81 $\pm$ 0.16& 1.09&21.06&16.41& 5.27&10.96 $\pm$ 0.12\\
CCDG-Reg				&11.01&17.61& 6.34& 5.95&10.23 $\pm$ 0.41& 1.10&21.86&16.59& 5.24&11.19 $\pm$ 0.18\\
CCDG-Ens				&11.47&17.07& 6.03& 5.92&10.12 $\pm$ 0.23& 1.11&21.37&16.60& 5.35&11.11 $\pm$ 0.15\\
\textbf{CTS (ours)}     & 4.56& 8.33& 3.24&	2.88& \textbf{4.75 $\pm$ 0.52}& 2.05&12.50& 9.68& 1.89& \textbf{6.53 $\pm$ 0.31}\\
\cmidrule{1-11}
TS with test      & 3.07& 4.26& 2.28& 2.59& 3.05 $\pm$ 0.31& 1.50& 6.04& 6.77& 4.64& 4.74 $\pm$ 0.11\\	
\bottomrule
\end{tabular}
\end{subtable}
\caption{Calibration errors on four different datasets. Each column represents the target domain in one-domain-leave-out setting.} \label{table:main1}
\end{table*}


\subsection{Experimental Settings}

\paragraph{Datasets.} We evaluate our CTS on four datasets that consist of multiple domains: PACS \cite{li2017deeper}, Office-Home \cite{venkateswara2017deep}, Digits-DG \cite{zhou2020learning} and VLCS \cite{fang2013unbiased}. PACS consists of 7 classes from 4 different domains and Office-Home has 65 classes from 4 domains. Digit-DG contains 10 classes from 4 domains and VLCS has 4 domains with 5 classes.

\paragraph{Performance metric.} 

As in previous works on calibration, we use expected calibration error (ECE) \cite{naeini2015obtaining} as a performance metric in our experiments. ECE measures the calibration error based on the expected absolute difference between the model's averaged confidence and its accuracy after binning.  By dividing the confidence score range $[0,1]$ into $R$ bins at even intervals $\text{(} q_r, q_{r+1} \text{]}$, we define the $r$-th bin $B_r$ as the indices of samples contained in that bin, i.e. $B_r = \{i=1 ... N~|~ \mathbb{P}(y=\hat{y}_i|\cdot) \in \text{(} q_r, q_{r+1} \text{]} \}$. Then, ECE can be calculated by $\sum_{r=1}^R \frac{|B_r|}{N}|Acc(B_r)-Conf(B_r)|$, where $Acc(B_r)=|B_r|^{-1} \sum_{i \in B_r} \mathds{1}(\hat{y}_i=y_i)$ represents the expected accuracy of samples in $B_r$  and $Conf(B_r)=|B_r|^{-1} \sum_{i \in B_r} \mathbb{P}(y=\hat{y}_i|\cdot)$ is the averaged confidence of $B_r$. Here, $\mathds{1}(A)$ is an indicator function with $\mathds{1}(A)=1$ if $A$ is true and $\mathds{1}(A)=0$, otherwise. $y_i$ is the ground truth class and $\hat{y}_i=\argmax_y \mathbb{P}(y|f_i,T)$ denotes the predicted class. We set $R=10$ during experiments. All results are averaged over 5 different runs and we report the results with 95\% confidence intervals.

\paragraph{Baselines.} We consider the following latest TS baselines: 1) \textit{Vanilla} \cite{hendrycks2017baseline}: The baseline without any calibration techniques applied; 2) \textit{TS with val/test} \cite{guo2017calibration}: TS with validation of the source domains/test set of target domain. The latter can be viewed as one of the ideal cases where the target domain information is given during calibration, which is impractical in many cases; 3) \textit{PerturbTS} \cite{tomani2021post}: TS by injecting some perturbations into the validation set; 4) \textit{CCDG} \cite{gong2021confidence}: Cluster-level TS with multiple calibration domains in  the validation set. We consider three variants of \textit{CCDG}. \textit{CCDG-NN} assigns each target sample to its nearest neighbor (NN) cluster's temperature. \textit{CCDG-Reg} employs regression-based mapping from sample-wise feature to the cluster-level temperature. \textit{CCDG-Ens} is a method of ensembles of \textit{TS with val}, \textit{CCDG-NN} and \textit{CCDG-Reg} in a logit space; 5) \textit{MDTS} \cite{yu2022robust}: Multi-domain TS with linear regression of temperatures.
Implementation of each method follows the corresponding paper and we leave the details   to the supplementary material. We also compare our method with train-time calibration methods in the supplementary material and show that CTS achieves comparable performance.

\paragraph{Implementation details.} 
As each dataset consists of four domains, we separate them into three source domains and one target domain following the one-domain-leave-out setup  \cite{zhou2021mixstyle, li2022uncertainty}. The three source domains are split again into training set and validation set following \cite{zhou2022domain, zhou2021domain22}, where we use the training set for training the model and the validation set for post-hoc calibration.  For a fair comparison, for all baselines except \textit{TS with test}, the OOD calibration performance is measured with the following process: (i) The model is first trained using the training set of source domains. (ii) Post-hoc calibration is conducted following each scheme using the validation set of source domains. (iii) ECE of each scheme is obtained using the target domain. 
During the training process in step (i), we utilize ResNet-18  \cite{he2016deep} pre-trained on ImageNet \cite{krizhevsky2017imagenet} and adopt the base training setup (without any DG techniques) following  \cite{zhou2021mixstyle} for all schemes, to mainly focus on the effect of post-hoc calibration.   Later in Fig. \ref{fig:plugin}, we also combine the baselines and our CTS with existing DG methods \cite{zhou2021mixstyle, li2022uncertainty}. For our CTS, the intermediate features for style and content shifts are extracted at the output of the first residual block of ResNet-18 in all experiments. Other details and ablation studies on the effect of the different  layer are also reported in the supplementary material.



\subsection{Experimental Results}

\paragraph{Comparison with post-hoc OOD calibration methods.} Tab. \ref{table:main1} compares the   ECEs of different schemes in each dataset. 
Since the post-hoc methods do not change the model parameters after calibration, all schemes have the same accuracy. In Tab.   \ref{table:main1}, we can see that our CTS achieves superior OOD calibration performance for all datasets compared to other post-hoc calibration methods. In particular, our method is much better for datasets with relatively large domain disparity, such as PACS and Digits-DG. This supports our claim that the temperature with style consistency can help achieve high calibration performance by regularizing the predictions for samples in the target domain, which has a large domain gap with the source domains. Also, for Office-Home dataset, when the coherent predictions within the same class are the matter due to the relatively large number of classes (65 classes in Office-Home), our CTS, which considers content consistency, can achieve the best performance by encouraging consistent predictions in the target domain. Since the model is pre-trained on ImageNet, we note that post-hoc calibration methods can negatively affect the calibration when the target domain is a photo in PACS dataset.

\begin{table}[!t]
\small
\centering
\begin{tabular}{c||c|c|c|c}
\toprule 
Methods    & PACS & Office-Home & Digits-DG & VLCS \\
\midrule
Vanilla    & 9.83&10.48&12.55&16.93\\
TS with val  & 9.06& 9.79&11.08&14.00\\
Class-wise & 6.27& 5.25& 6.96&13.70\\
CTS (only S)      & 5.09& 5.51& 6.82&13.10\\
CTS (only C)      & 5.47& 5.36& 7.13&13.19\\
\textbf{CTS (S\&C)}    & \textbf{4.31}& \textbf{4.75}& \textbf{6.53}& \textbf{12.90}\\
\bottomrule
\end{tabular}
\caption{Ablation studies.  We report the  calibration errors for different variations of CTS: (i) class-wise, where the difference between confidence scores obtained from two samples within the same class is minimized; (ii) only S or only C, where either the style or content is solely considered instead of using both of them. The results demonstrate the effectiveness of each component in CTS with meaningful error gaps.} \label{table:ablation1}
\end{table}

\paragraph{Ablations for our CTS.}  One key feature of our CTS is to  balance between style and content consistencies by adjusting coefficients $\lambda_1$ and $\lambda_2$ in Eq. \ref{totalloss}. This enables CTS to control the trade-off between two consistencies depending on the degree of disparity in terms of styles and contents, which may vary depending on the dataset. In Tab.   \ref{table:ablation1}, we conduct ablation studies on CTS by considering different variations of our method. One variation we can think of is to directly minimize  the difference between confidence scores obtained from two samples within the same category, in a class-wise manner (i.e., class-wise in Tab. \ref{table:ablation1}). Interestingly, the class-wise method achieves a relatively good calibration performance compared to other baselines (e.g., TS with val), which substantiates our claim that consistency-guided temperature is a good solution for OOD calibration.  We also consider adopting either the style consistency loss or content consistency loss solely (i.e., only S or  only C  in Tab.   \ref{table:ablation1}), instead of  using both of them. One interesting observation is that style-consistency (i.e., only S) has more advantages in datasets such as PACS or Digits-DG where styles are relatively the main cause of disparity compared to contents, while content-consistency (i.e., only C) performs better when contents become more important as in Office-Home. Our CTS considering both styles and contents can strategically balance between them depending on the dataset, taking the best of both worlds.

\begin{figure}[t]
\centering
\begin{subfigure}[b]{0.23\textwidth}
\centering
\includegraphics[width=\textwidth]{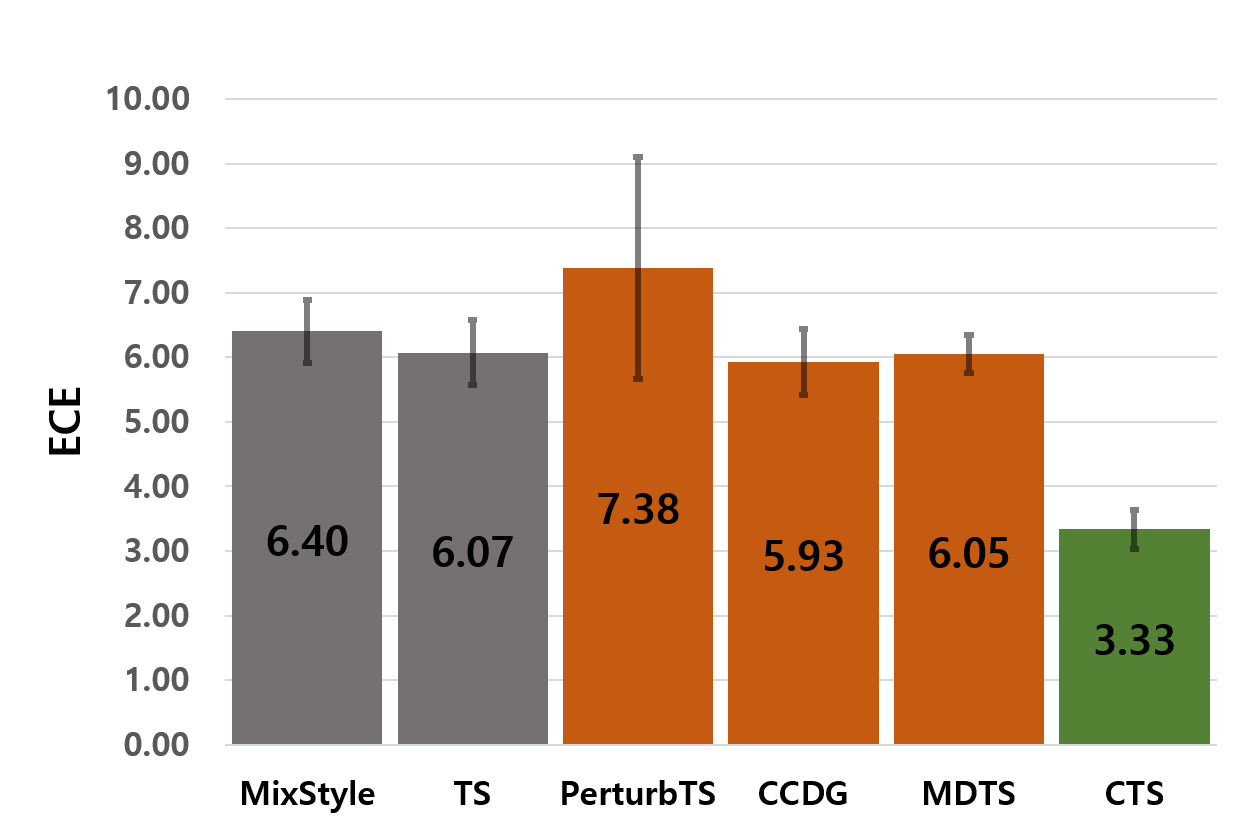}
\caption{Compatibility with MixStyle}
\label{fig:aaaaaaaaa}    
\end{subfigure}
\begin{subfigure}[b]{0.23\textwidth}
\centering
\includegraphics[width=\textwidth]{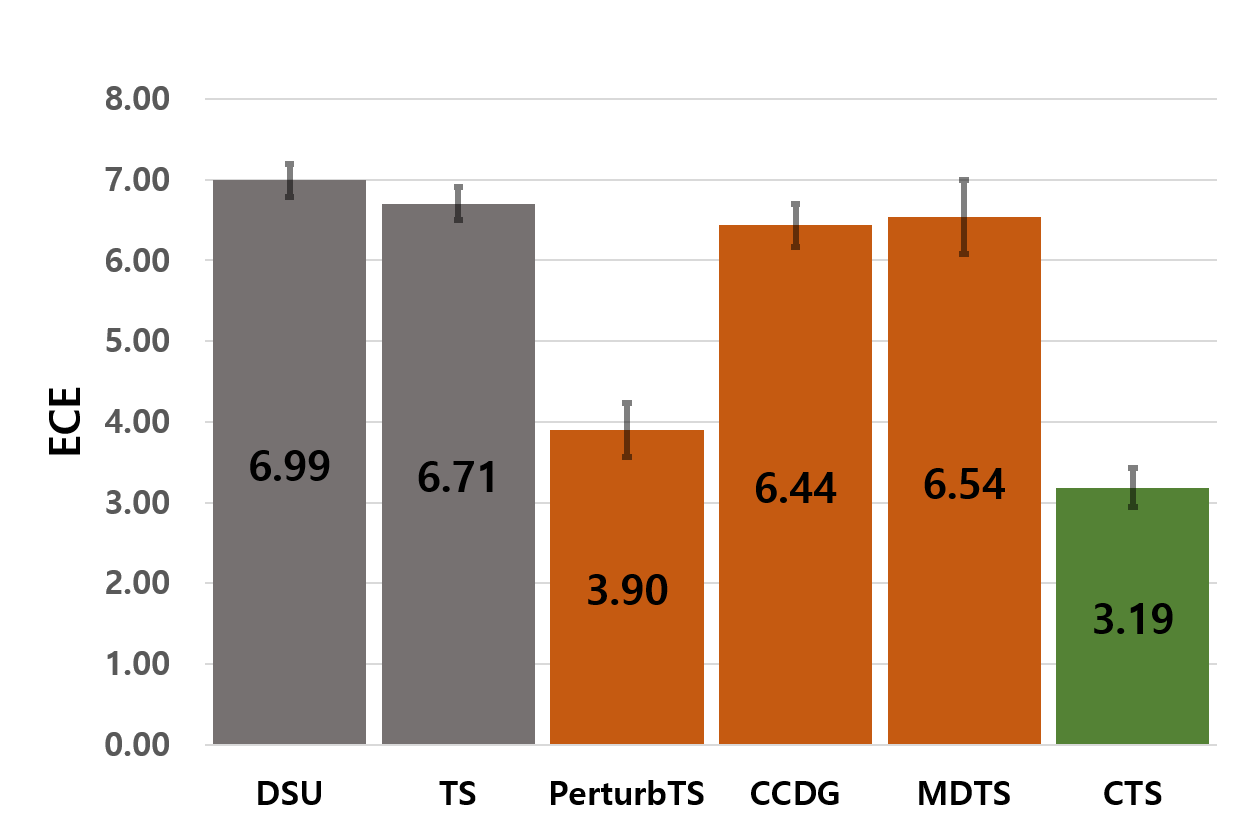}
\caption{Compatibility with DSU}
\label{fig:bbbbbbbbbb}    
\end{subfigure}
\caption{Compatibility with augmentation-based DG methods on PACS dataset.  Each scheme is combined with either MixStyle \cite{zhou2021mixstyle} or DSU \cite{li2022uncertainty}.}  \label{fig:plugin}
\end{figure}

\paragraph{Compatibility with existing DG methods.} Since our CTS is a post-hoc calibration method based on TS, it has the advantage of being easily combined with existing domain generalization methods without compromising accuracy. To demonstrate this, we measure the OOD calibration performance by applying calibration methods to MixStyle \cite{zhou2021mixstyle} and DSU \cite{li2022uncertainty}, which are the recently proposed augmentation-based DG approaches. As shown in Fig. \ref{fig:plugin}, our CTS significantly improves the calibration performance of DG methods compared to other baselines. These results pave the way to apply existing DG methods in trustworthy AI systems, further confirming the advantage of CTS.


%

\paragraph{Effect of domain compositions in train/validations sets.}  

In Tab. \ref{table:composition}, we also report the OOD calibration results of multi-domain calibration schemes when the training and validation sets have separated domain compositions without overlapping as in \cite{gong2021confidence, yu2022robust}: One domain is used for training, two are utilized for post-hoc calibration, and the remaining one is used to measure the ECE and accuracy.  PACS dataset is utilized for experiments. In this setup, the accuracy of each scheme decreases from 77.95\% to 50.2\% since the model is trained  with only one source domain. 
Interestingly, despite the decrease in the number of calibration domains, our CTS still achieves higher OOD calibration performance than other baselines. These results show that CTS can effectively utilize  the intrinsic attributes of validation samples to encourage consistent predictions regardless of style and content variations, 
even when the number of domains for calibration is limited.




\begin{table}[!t]
\addtolength{\tabcolsep}{-4.5pt}
\small
\centering
\begin{tabular}{l||l||cccc|c|c}
\toprule  
\makecell{\# of cali.\\ domains} &  Methods   & Art & Cartoon & Photo & Sketch & Avg. & Acc \\
\midrule
& CCDG-NN               &10.36&25.75& 6.97&10.52&13.40&50.20 \\	
&CCDG-Reg              &11.33&22.96& 5.97&12.82&13.27&50.20 \\	
2     & CCDG-Ens              &11.00&22.15& 6.86&11.89&12.98&50.20 \\	
&MDTS                  &14.80&20.11& 8.27&12.10&13.82&50.20 \\	
&\textbf{CTS}  & 9.91&14.59&11.01&12.84&\textbf{12.09}&50.20 \\	
\midrule
3    &\textbf{CTS} & 3.20& 3.05& 8.59& 2.38&\textbf{ 4.31}&77.95 \\	
\bottomrule
\end{tabular}
\caption{Effect of domain  composition in multi-domain calibration schemes. Our CTS still achieves higher OOD calibration performance than others, confirming the effectiveness of our consistency-guided approach.} \label{table:composition}
\end{table}

\paragraph{Additional experimental results.}  Other experimental results including large-scale dataset, coefficient ablation ($\lambda_1, \lambda_2$), layer ablation, and comparison with train-time calibration methods 
are reported in the supplementary material.

\section{Conclusion}
In this work, we proposed consistency-guided temperature scaling (CTS), a post-hoc calibration method that can be effectively used for robust OOD predictions. Our CTS strategically utilizes the style and content information of the source domains in a multi-domain setting, encouraging consistent predictions under domain shifts. By introducing auxiliary losses that can take consistencies into account in terms of two key aspects,  styles and contents, our CTS can optimize the scaling temperature targeting OOD settings.  Our approach achieves superior OOD  calibration performance  without requiring any target domain information and without compromising the model accuracy,   making the scheme directly applicable to various trustworthy AI systems. 


\section*{Acknowledgments}
This work was supported by IITP funds from MSIT of Korea (No. 2020-0-00626), NRF (No. 2019R1I1A2A02061135). 
Dong-Jun Han is the corresponding author.


\bibliography{aaai24}

\end{document}